\documentclass[runningheads]{llncs}
\usepackage{graphicx}
\usepackage{amsmath,amssymb} % define this before the line numbering.
\usepackage{color}

\usepackage{algorithm,algpseudocode}
\usepackage{hyperref}
\begin{document}
\pagestyle{headings}
\mainmatter

% \def\ACCV18SubNumber{}  % Insert your submission number here

%===========================================================
\title{Compact retail shelf segmentation for mobile deployment} % Replace with your title
% \titlerunning{ACCV-18 submission ID \ACCV18SubNumber}
% \authorrunning{ACCV-18 submission ID \ACCV18SubNumber}

\author{Pratyush Kumar, Muktabh Mayank Srivastava}
\institute{ParallelDots, Inc. \thanks{https://www.paralleldots.com} \\
\email{\{pratyush,muktabh\}@paralleldots.com}}

% \thanks{*paralleldots.xyz}

\maketitle

%===========================================================
\begin{abstract}
The recent surge of automation in the retail industries has rapidly increased demand for applying deep learning models on mobile devices. To make the deep learning models real time on-device, a compact efficient network becomes inevitable. In this paper, we work on one such common problem in the retail industries - \emph{Shelf segmentation}. Shelf segmentation can be interpreted as a pixel-wise classification problem, i.e., each pixel is classified as to whether they belong to visible shelf edges or not. The aim is not just to segment shelf edges, but also to deploy the model on mobile devices. As there is no standard solution for such dense classification problem on mobile devices, we look at semantic segmentation architectures which can be deployed on edge. We modify low-footprint semantic segmentation architectures to perform shelf segmentation. In addressing this issue, we modified the famous U-net architecture in certain aspects to make it fit for on-devices without impacting significant drop in accuracy and also with 15X fewer parameters. In this paper, we proposed Light Weight Segmentation Network (LWSNet), a small compact model able to run fast on devices with limited memory and can train with less amount ($\sim$ 100) of labeled data.
% \dots
\end{abstract}

% As there is no standard solution for this problem, we used semantic segmentation approaches and tune them to make it suitable for our problem. Also, current state-of-the-art segmentation networks have enormous amounts of parameters which lead to large model size hence unsuitable for deploying on mobile devices. To solve this issue, a commonly exploited workaround is to reduce the network size while ignoring the inherent characteristics of semantic segmentation, but then deep networks show a critical performance drop.

\section{Introduction}

In the last few years, deep convolutional neural networks (CNNs) \cite{LecunGradient-BasedRecognition} have outperformed the state of the art in many visual recognition tasks. CNNs have been responsible for the phenomenal advancements in tasks like object classification \cite{KrizhevskyImageNetNetworks}, object localization \cite{Sermanet2013OverFeat:Networks} etc., and the continuous improvements to CNN architectures are bringing further progress \cite{Simonyan2014VeryRecognition,Szegedy2014GoingConvolutions}. In many visual tasks, especially in retail industries, the desired output should include localization (or semantic segmentation), i.e., a class label is supposed to be assigned to each pixel or pixel-wise classification. Development of recent deep convolutional neural networks (CNNs) makes remarkable progress on semantic segmentation \cite{Chen2014SemanticCRFs,Chen2018SearchingPrediction,Chen2018Encoder-DecoderSegmentation}. The effectiveness of these networks largely depends on the sophisticated model design regarding depth and width, which has to involve many operations and parameters. The CNN based semantic segmentation has been widely used in different applications with great accuracy.\\*

Recent interest in the automation of retail and consumer industries has a strong demand for deploying shelf segmentation models on mobile devices. Few such applications are 1) to determine whether a photo is being taken from front or corner while doing shelf eye tracking studies. Eye tracking devices are used to measure where a person is looking, also known as our point of gaze. These measurements are carried out by an eye tracker, that records the position of the eyes and the movements they make. 2) measuring retail execution or in-store execution. Retail execution aims to put the right products are in the right place on the retail shelf with the right amount of space between them. 3) measuring planogram compliance. A planogram is a visual representation that shows how and where specific retail products should be placed on retail shelves. Planogram compliance is part of the retail execution and it ensures stores follow the rules which capture optimal consumer attention. 4) separating shelves to do an out of stocks analysis. To solve these problems on a mobile device, there are no standard architectures and real baselines, we thus begin with use of standard deep architectures for dense pixel-wise classification on shelf images and adapt them for accuracy and deployment on mobile devices. There are some architectures proposed for low resource semantic segmentation which are used as a baseline. \\*

\begin{figure}[tb]
  \centering
  \includegraphics[width=\textwidth]{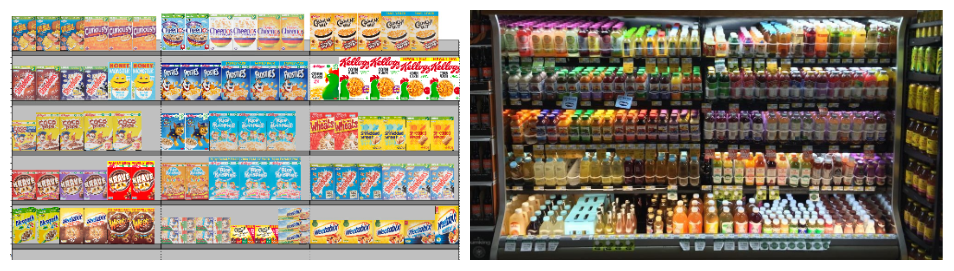}
  \caption{A planogram (left side) visual representation. Retail execution (right side) showing products arranged systematically ensuring planogram compliance.}
  \label{fig:unet}
\end{figure}

It is challenging to design a model with both small memory footprint and high accuracy. All the current state-of-the-art approaches exploit deep learning architectures and are typically based on fully convolutional networks \cite{Long2014FullySegmentation} with an encoder-decoder architecture. Deep neural networks allow to obtain impressive performance, however due to millions of parameters the model size is huge which makes their deployment on mobile very inefficient. So, efficient neural networks are becoming ubiquitous in mobile applications enabling entirely new on-device experiences.\\*

CNNs have shown a great result for various semantic segmentation tasks in recent years. There has also been research to find low resource semantic segmentation architectures, unlike for Dense Pixelwise Classification. As there is no standard solution for shelf segmentation on phones, we used semantic segmentation approaches as a baseline for the task. U-Net \cite{RonnebergerU-Net:Segmentation} is a famous architecture for dense classification. The structure of U-Net, comprising an encoder and a decoder network, and the corresponding layers of the encoder and decoder network are connected by skip connection, prior to a pooling and subsequent to a deconvolution operation respectively. U-Net has been showing impressive potential in segmenting images, even with a scarce amount of labeled training data. \\*

The proposed LWSNet is in the direction to cover the limitations for mobile deployment, which is a small compact model able to run fast on devices with limited memory.

% In this paper, we modified the U-Net architecture and added diligently some different perspective and famous approach to capture an image details without going much deeper architecture. We mainly focused on reducing depth and enhancing breadth of the model. \\*

\section{Related Work}
Different classes of deep learning based dense classification architectures have been proposed. Most of the models follow the fully convolutional networks approach with an Encoder-Decoder type of architecture. In the encoder part of these networks, the feature extractors are powerful object detectors like ResNet \cite{He2015DeepRecognition}, ResNext \cite{Xie2016AggregatedNetworks}, etc. The most famous encoder-decoder network is U-Net which gives good accuracy with few training data. \\*

Designing deep neural network architecture for the optimal trade-off between accuracy and efficiency has been an active research area in recent years. SqueezeNet \cite{Iandola2016SqueezeNet:Size} extensively uses 1x1 convolutions with squeeze and expand modules primarily focusing on reducing the number of parameters. In this section, we introduce related work on small segmentation models. Small semantic segmentation models require making a good trade-off between accuracy and model parameters. More recent works that focus on real-time efficient segmentation are ERFNet \cite{RomeraERFNet:Segmentation}, ENet \cite{Paszke2016ENet:Segmentation}, ICNet \cite{Zhao2017ICNetImages}. However, most of them follow the design principles of image classification, which makes them have poor segmentation accuracy. \\*

Quantization \cite{Jacob2017QuantizationInference,Krishnamoorthi2018QuantizingWhitepaper,Wu2015QuantizedDevices} is another important complementary effort to improve network efficiency through reduced precision arithmetic. Post-training quantization reduces model size while improving CPU and hardware accelerator latency, with little degradation in model accuracy. There are many quantization techniques e.g., dynamic range quantization, full integer quantization, float16 quantization. \\*

% Our work is more focused on using efficient CNN modules and better training procedures by keeping the macro architecture remain same. In this paper, we adopt the U-Net macro architecture and experiments with convolution module, model compression and design space.  

\section{Motivation and building blocks}

The main focus of our work is to use encoder-decoder architecture with skip connections and efficient CNN modules by keeping the macro structure remain same. We adopt the U-Net macro architecture and experiments with convolution module, model compression and design space.

\begin{figure}[tb]
  \centering
  \includegraphics[width=\textwidth]{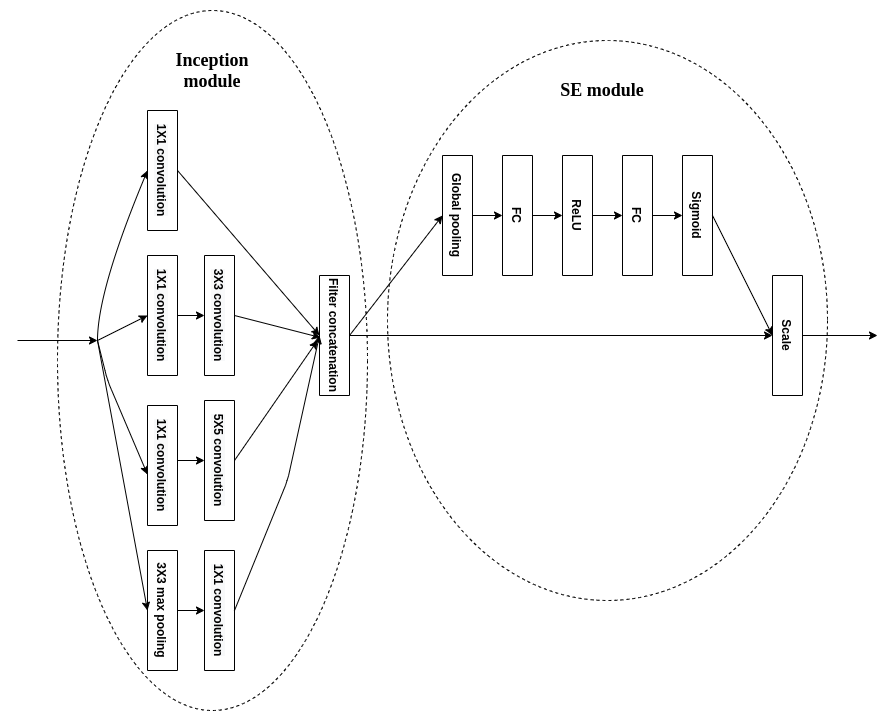}
  \caption{The broad schema of Inception-SE module}
  \label{fig:zoom}
\end{figure}

\subsection{U-Net}
The network architecture is symmetric, having an Encoder that extracts spatial features from the image, and a Decoder that constructs the segmentation map from the encoded features. The Encoder follows the typical formation of a convolutional network. It involves a sequence of convolution operations, followed by a max pooling operation. The Decoder part up-samples the feature map using a transposed convolution operation and reduces the feature channels, followed by a sequence of convolution operations. The most ingenious aspect of the U-Net architecture is the introduction of skip connections. In every layer, the output of the convolutional layer of the Encoder is transferred to the Decoder.

\subsection{Inception module}
Inception \cite{Szegedy2014GoingConvolutions} modules are used in Convolutional Neural Networks to allow for more efficient computation and deeper Networks through dimensionality reduction with stacked 1x1 convolutions. The modules were designed to solve the problem of computational expense, as well as overfitting, among other issues. The solution, in short, is to take multiple kernel filter sizes within the CNN, and rather than stacking them sequentially, ordering them to operate on the same level. Inception Layer is a combination of all those layers (namely, 1x1 Convolutional layer, 3x3 Convolutional layer, 5x5 Convolutional layer) with their output filter banks concatenated into a single output vector forming the input of the next stage.

\begin{figure}[tb]
  \centering
  \includegraphics[width=\textwidth]{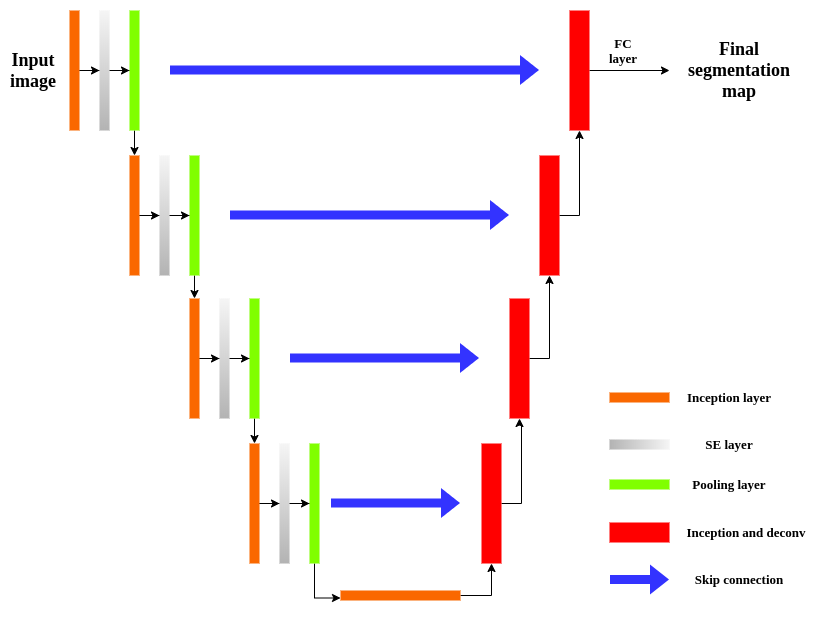}
  \caption{The U-Net architecture with Inception-SE module}
  \label{fig:unet}
\end{figure}

\subsection{Squeeze and Excitation (SE) module}
The main goal of the SE \cite{Hu2017Squeeze-and-ExcitationNetworks} module is to make a good trade-off between improved performance and increased model complexity. The SE is a small network that tries to detect a good pattern in global average of features, and then excites or suppresses those features in a way that helps classification. While it should be noted that the SE modules themselves add depth, they do so in an extremely computationally efficient manner and yield good returns even at the point at which extending the depth of the base architecture achieves diminishing returns. It has been seen that the gains are consistent across a range of different network depths, suggesting that the improvements induced by SE modules may be complementary to those obtained by simply increasing the depth of the base architecture. 

\section{Proposed architecture}

The network architecture is illustrated in Fig. ~\ref{fig:unet}. It consists of a contracting path or encoder (left side) and an expansive path or decoder (right side). The encoder consists of the repeated application of Inception layer, each followed by SE layer and a 2x2 max pooling operation with stride 2 for downsampling. At each downsampling step we double the number of feature channels. Every step in the decoder path consists of an upsampling of the feature map followed by a 2x2 up-convolution that halves the number of feature channels, a concatenation with corresponding cropped feature from the encoder path and Inception layer. At the final layer a 1x1 convolution is used to map each final feature vector to the desired number of classes. We used \# of filters = [32, 64, 128, 256, 512] along the levels. 

\setlength{\tabcolsep}{8pt}
\begin{table}[H]
\begin{center}
\caption{LWSNet architecture}
\label{table:result1}
\begin{tabular}{lll}
\hline\noalign{\smallskip}
Input name/type & Output size & \# of parameters\\
\noalign{\smallskip}
\hline
\noalign{\smallskip}
Input image & 1x224x224 & -\\
Inception-1 & 32x224x224 & 2,100\\
SE-1 & 32x224x224 & 162\\
MaxPool-1 & 32x112x112 & -\\
Inception-2 & 64x112x112 & 9,644\\
SE-2 & 64x112x112 & 580\\
MaxPool-2 & 64x56x56 & -\\
Inception-3 & 128x56x56 & 38,056\\
SE-3 & 128x56x56 & 2,184\\
MaxPool-3 & 128x28x28 & -\\
Inception-4 & 256x28x28 & 151,120\\
SE-4 & 256x28x28 & 8,464\\
MaxPool-4 & 256x14x14 & -\\
Inception-5 & 512x14x14 & 602,272\\
Deconv-1 & 256x28x28 & 524,544\\
Inception-6 & 256x28x28 & 230,992\\
Deconv-2 & 128x56x56 & 131,200\\
Inception-7 & 128x56x56 & 58,024\\
Deconv-3 & 64x112x112 & 32,832\\
Inception-8 & 64x112x112 & 14,644\\
Deconv-4 & 32x224x224 & 8,224\\
Inception-9 & 32x224x224 & 3,730\\
Final conv & 2x224x224 & 66\\
\hline
\end{tabular}
\end{center}
\end{table}
\setlength{\tabcolsep}{1.4pt}

\subsection{Comparison of models}

We compare the number of parameters for each layer in Table ~\ref{table:result1}. In the contraction path, there are 5 Inception layer, 4 SE layer, each followed by MaxPool. In the expansion path, there is a deconv layer, concatenated with the corresponding feature path from the contracting path, each followed by the Inception layer. There are 4 deconv layer and 4 Inception layer. At the end, 1x1 convolutional layer to the final number of classes. In Table ~\ref{table:result2}, LWSNet uses around $\sim$ 1.8 million parameters compare to $\sim$ 2.2 million and $\sim$ 22 million in ERFNet and ICNet respectively. There is a significant reduction in the parameters. We used post-training quantization when converting the LWSNet model to TensorFlow Lite format. This further has helped to reduce the model size by 3X, with the final model weight $\sim$ 7MB. In the result section, you see there is a slight improvement in accuracy compared to both these models. \\*

\setlength{\tabcolsep}{8pt}
\begin{table}[H]
\begin{center}
\caption{Number of parameters for different architectures}
\label{table:result2}
\begin{tabular}{lll}
\hline\noalign{\smallskip}
Model & Number of parameters & Model size (MB)\\
\noalign{\smallskip}
\hline
\noalign{\smallskip}
LWSNet & 1,818,838 & 22.0\\
ERFNet \cite{RomeraERFNet:Segmentation} & 2,063,086 & 24.9\\
ICNet \cite{Zhao2017ICNetImages} & 22,726,752 & 140.7\\
\hline
\end{tabular}
\end{center}
\end{table}
\setlength{\tabcolsep}{1.4pt}

\section{Datasets}
\begin{figure}[tb]
  \centering
  \includegraphics[width=\textwidth]{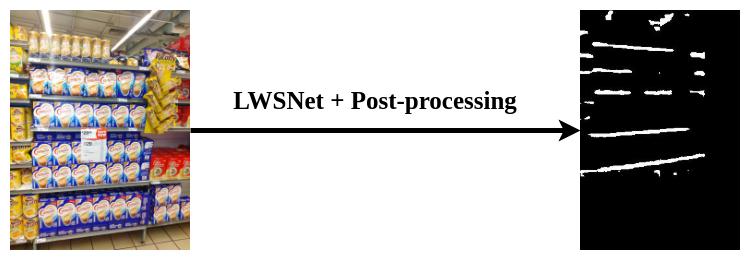}
  \caption{Original image (left) and prediction through the model (right)}
  \label{fig:output}
\end{figure}

We used our in-house datasets of retail industries for shelf segmentation. In our training datasets, we have 98 training images of shelve and products in the stores. These all 98 training images come from one domain set. We used 100 test images from 4 domains e.g., test set 1, 2, 3, 4; with each domain containing 25 images. The training and test set 1 have images similar to one shown in Fig. ~\ref{fig:output}. In this domain, shelve images are taken closely and each image has one complete block of a shelve. In test set 2, images of shelve are taken from distant. Similarly in test set 3, there is more than one complete block of shelve and in test set 4, there are both distant images and more than one complete block of shelve. The training set contains images of different scales and sizes, ranging from 400 pixels to 3000 pixels. We cropped high resolution images to multiple windows so that the model can learn low level information more efficiently. After LWSNet model prediction, we use that output and do post-processing to complete shelf edges segmented as shown in Fig. ~\ref{fig:output}.

\section{Results}
We have tried the model for 4 test sets with 25 images each and each set belongs to a different domain. We calculated Jaccard Index (IoU) of the LWSNet with two similar state-of-the-art dense classification models for on-device, ERFNet and ICNet. Our main target focused mainly to make a good trade-off between the number of parameters and accuracy. In our in-house datasets, our approach lead to give improved accuracy with large reduction in the model size.

\setlength{\tabcolsep}{8pt}
\begin{table}[H]
\begin{center}
\caption{Accuracies of different models on multiple test sets.}
\label{table:result3}
\begin{tabular}{lllll}
\hline\noalign{\smallskip}
Model & Test set 1 & Test set 2 & Test set 3 & Test set 4\\
\noalign{\smallskip}
\hline
\noalign{\smallskip}
LWSNet & \textbf{0.458} & \textbf{0.445} & \textbf{0.449} & \textbf{0.439}\\
ICNet \cite{Zhao2017ICNetImages} & 0.448 & 0.441 & 0.445 & 0.433\\
ERFNet \cite{RomeraERFNet:Segmentation} & 0.439 & 0.432 & 0.429 & 0.424\\
\hline
\end{tabular}
\end{center}
\end{table}
\setlength{\tabcolsep}{1.4pt}

\section{Conclusion}
The LWSNet has performed better in terms of accuracy and with lesser numbers of parameters. We beat the state-of-the-art networks of similar types of models on our in-house datasets. Shelf segmentation is a common problem in the retail industries but has broad scope to solve it with deep neural networks. In the future work, more research will be required to solve it more efficiently but encoder-decoder architecture indeed would be taken as baseline. 

\section{Acknowledgments}
The authors would like to thank Srikrishna Varadarajan, Sonaal Kant, and Harshita Seth for their contributions in gathering datasets and helping with the post-processing stage.

\bibliographystyle{splncs}
\bibliography{references}

\begin{thebibliography}{10}

\bibitem{LecunGradient-BasedRecognition}
Lecun, Y., Eon~Bottou, L., Bengio, Y., Abstract|, P.H.:
\newblock {Gradient-Based Learning Applied to Document Recognition}.
\newblock (Technical report)

\bibitem{KrizhevskyImageNetNetworks}
Krizhevsky, A., Sutskever, I., Hinton, G.E.:
\newblock {ImageNet Classification with Deep Convolutional Neural Networks}.
\newblock (Technical report)

\bibitem{Sermanet2013OverFeat:Networks}
Sermanet, P., Eigen, D., Zhang, X., Mathieu, M., Fergus, R., LeCun, Y.:
\newblock {OverFeat: Integrated Recognition, Localization and Detection using
  Convolutional Networks}.
\newblock (2013)

\bibitem{Simonyan2014VeryRecognition}
Simonyan, K., Zisserman, A.:
\newblock {Very Deep Convolutional Networks for Large-Scale Image Recognition}.
\newblock (2014)

\bibitem{Szegedy2014GoingConvolutions}
Szegedy, C., Liu, W., Jia, Y., Sermanet, P., Reed, S., Anguelov, D., Erhan, D.,
  Vanhoucke, V., Rabinovich, A.:
\newblock {Going Deeper with Convolutions}.
\newblock (2014)

\bibitem{Chen2014SemanticCRFs}
Chen, L.C., Papandreou, G., Kokkinos, I., Murphy, K., Yuille, A.L.:
\newblock {Semantic Image Segmentation with Deep Convolutional Nets and Fully
  Connected CRFs}.
\newblock (2014)

\bibitem{Chen2018SearchingPrediction}
Chen, L.C., Collins, M.D., Zhu, Y., Papandreou, G., Zoph, B., Schroff, F.,
  Adam, H., Shlens, J.:
\newblock {Searching for Efficient Multi-Scale Architectures for Dense Image
  Prediction}.
\newblock (2018)

\bibitem{Chen2018Encoder-DecoderSegmentation}
Chen, L.C., Zhu, Y., Papandreou, G., Schroff, F., Adam, H.:
\newblock {Encoder-Decoder with Atrous Separable Convolution for Semantic Image
  Segmentation}.
\newblock (2018)

\bibitem{Long2014FullySegmentation}
Long, J., Shelhamer, E., Darrell, T.:
\newblock {Fully Convolutional Networks for Semantic Segmentation}.
\newblock (2014)

\bibitem{RonnebergerU-Net:Segmentation}
Ronneberger, O., Fischer, P., Brox, T.:
\newblock
\newblock ({U-Net: Convolutional Networks for Biomedical Image Segmentation})

\bibitem{He2015DeepRecognition}
He, K., Zhang, X., Ren, S., Sun, J.:
\newblock {Deep Residual Learning for Image Recognition}.
\newblock (2015)

\bibitem{Xie2016AggregatedNetworks}
Xie, S., Girshick, R., Doll{\'{a}}r, P., Tu, Z., He, K.:
\newblock {Aggregated Residual Transformations for Deep Neural Networks}.
\newblock (2016)

\bibitem{Iandola2016SqueezeNet:Size}
Iandola, F.N., Han, S., Moskewicz, M.W., Ashraf, K., Dally, W.J., Keutzer, K.:
\newblock {SqueezeNet: AlexNet-level accuracy with 50x fewer parameters and
  <0.5MB model size}.
\newblock (2016)

\bibitem{RomeraERFNet:Segmentation}
Romera, E., Alvarez, J.M., Bergasa, L.M., Arroyo, R.:
\newblock {ERFNet: Efficient Residual Factorized ConvNet for Real-time Semantic
  Segmentation}.
\newblock (Technical report)

\bibitem{Paszke2016ENet:Segmentation}
Paszke, A., Chaurasia, A., Kim, S., Culurciello, E.:
\newblock {ENet: A Deep Neural Network Architecture for Real-Time Semantic
  Segmentation}.
\newblock (2016)

\bibitem{Zhao2017ICNetImages}
Zhao, H., Qi, X., Shen, X., Shi, J., Jia, J.:
\newblock {ICNet for Real-Time Semantic Segmentation on High-Resolution
  Images}.
\newblock (2017)

\bibitem{Jacob2017QuantizationInference}
Jacob, B., Kligys, S., Chen, B., Zhu, M., Tang, M., Howard, A., Adam, H.,
  Kalenichenko, D.:
\newblock {Quantization and Training of Neural Networks for Efficient
  Integer-Arithmetic-Only Inference}.
\newblock (2017)

\bibitem{Krishnamoorthi2018QuantizingWhitepaper}
Krishnamoorthi, R.:
\newblock {Quantizing deep convolutional networks for efficient inference: A
  whitepaper}.
\newblock (2018)

\bibitem{Wu2015QuantizedDevices}
Wu, J., Leng, C., Wang, Y., Hu, Q., Cheng, J.:
\newblock {Quantized Convolutional Neural Networks for Mobile Devices}.
\newblock (2015)

\bibitem{Hu2017Squeeze-and-ExcitationNetworks}
Hu, J., Shen, L., Albanie, S., Sun, G., Wu, E.:
\newblock {Squeeze-and-Excitation Networks}.
\newblock (2017)

\end{thebibliography}

\end{document}